\documentclass[english]{article}
\usepackage{eucal}
\usepackage[utf8]{inputenc}
\usepackage[T1]{fontenc}
\usepackage{babel}
\usepackage{amsmath}
\usepackage{amsfonts}
\usepackage{csquotes}
\usepackage{amstext}
\usepackage{amssymb}
\usepackage{graphicx}
\usepackage{comment}
\usepackage{subcaption}
\usepackage{fancyhdr}
\usepackage{siunitx}
\usepackage{hyperref}
\hypersetup{
    colorlinks=true,
    linkcolor=blue,
    filecolor=magenta,      
    urlcolor=cyan,
}
\usepackage{graphicx}

\usepackage{biblatex}
\begin{document}


\title{Deep reinforcement learning with automated label extraction from clinical reports accurately classifies 3D MRI brain volumes}

\author{ \textbf{Joseph Stember}$^1$
\and 
\textbf{Hrithwik Shalu}$^2$}
\maketitle
\thispagestyle{fancy}
\noindent
\textsuperscript{1}Memorial Sloan Kettering Cancer Center, New York, NY, US, 10065 
\\
\textsuperscript{2}Indian Institute of Technology, Madras, Chennai, India, 600036
\\
\noindent
\textsuperscript{1}joestember@gmail.com
\\
\textsuperscript{2}lucasprimesaiyan@gmail.com 
\\


\begin{abstract}

\indent \textit{Purpose} Image classification is perhaps the most fundamental task in imaging artificial intelligence. However, labeling images is time-consuming and tedious. We have recently demonstrated that reinforcement learning (RL) can classify 2D slices of MRI brain images with high accuracy. Here we make two important steps toward speeding image classification: Firstly, we automatically extract class labels from the clinical reports. Secondly, we extend our prior 2D classification work to fully 3D image volumes from our institution. Hence, we proceed as follows: in Part 1, we extract labels from reports automatically using a natural language processing approach termed sentence bidirectional encoder representations from transformers (SBERT). Then, in Part 2, we use these labels with RL to train a classification Deep-Q Network (DQN) for 3D image volumes.

\indent \textit{Materials and Methods} For Part 1, we trained SBERT with 45 "normal" patient report impressions and 45 metastasis-containing impressions. We then used the trained SBERT to predict class labels for use in Part 2. In Part 2, we applied multi-step image classification to allow for combined Deep-Q learning using 3D convolutions and TD(0) Q learning. We trained on a set of $90$ images ($40$ normal and $50$ tumor-containing). We tested on a separate set of $61$ images ($40$ normal and $21$ tumor-containing), again using the classes predicted from patient reports by the trained SBERT in Part 1. For comparison, we also trained and tested a supervised deep learning classification network on the same set of training and testing images using the same labels. 

\indent \textit{Results}
Part 1: Upon training with the corpus of radiology reports, the SBERT model had $100 \%$ accuracy for both normal and metastasis-containing scans. Part 2: Then, using these labels, whereas the supervised approach quickly overfit the training data and as expected performed poorly on the testing set ($ 66 \%$ accuracy, just over random guessing), the reinforcement learning approach achieved an accuracy of  $92 \%$. The results were found to be statistically significant, with a  $p$-value of $3.1 \times 10^{-5}$. 

\indent \textit{Conclusion}
We have shown proof-of-principle application of automated label extraction from radiological reports. Using these labels, we have built on prior work applying RL to classification, extending from 2D slices to full 3D image volumes. RL has again demonstrated a remarkable ability to train effectively, in a generalized manner, and based on very small training sets.

\end{abstract}

\pagebreak

\section*{Introduction}

Classification is widely recognized as one of if not the most important and fundamental tasks in radiology artificial intelligence (AI) \cite{mcbee2018deep,saba2019present,mazurowski2019deep}. As is the case for the vast majority of radiology AI tasks, the literature on AI radiology classification is dominated almost exclusively by supervised deep learning (SDL). SDL entails an expert user / radiologist manually annotating each of a large set of images with class labels, typically requiring hundreds or thousands of annotated images for adequate training, often also necessitating data augmentation to artificially increased the number of images. However, labeling images, particularly so many, is time-consuming and tedious. 

To summarize, two key hindrances in SDL are the following requirements:
\begin{enumerate}
    \item Hand annotation by researchers with domain knowledge, usually diagnostic radiologists.
    \item Retrieving and annotating large numbers of images, typically hundreds to thousands, often also requiring data augmentation.
\end{enumerate}

We have in recent work, employed reinforcement learning (RL) to address the second item for 2D images \cite{stember2021deep}. In this work we reduced the number of training images required by an order of magnitude or two, to the tens of images. Training on a mere 30 training set images, RL achieved perfect accuracy when applied to a separate testing set of 30 images \cite{stember2021deep}.

However, for realistic clinical deployment, we wish to apply classification to full 3D image volumes. Further, we would like to address item one from above; our 2D proof-of-principle still required radiologist manual annotation of class labels, although greatly reduced in number to the order of tens. If we could avoid the need for manual annotation, this would greatly speed research and allow for new classification models to be trained quickly as the need arises, either for clinical deployment or for research applications. 

In this work we address both of the above key challenges in AI image classification, doing so for full 3D image volumes. In Part 1, we extract image labels automatically from the radiology reports using attention-based natural language processing (NLP). In part 2, we use the labels generated from Part 1 to train an RL-based classifier for 3D image volumes of patients who have non-central nervous system primary cancer with either normal MRI brain scans (in the sense of no intracranial lesions) or with scans showing metastatic spread to the brain. 

\section*{Methods}

\subsection*{Data collection}

We obtained a waiver of informed consent and approval from the Memorial Sloan Kettering Cancer Center (MSKCC) Institutional Review Board. After identifying candidate studies through a search of MSKCC radiology archives (mPower,  Nuance Communications, Inc.) of MRI brain with and without contrast studies performed between January 1st, 2012 and December 30th, 2018, we extracted two data sets of MRI brain scans, including the reports in a spreadsheet. The extracted data included reports generated by 18 attending-level neuroradiologists at MSKCC. Exclusion criteria included:
\begin{itemize}
    \item T1-post-contrast images not 5 mm in slice thickness
    \item Non-metastatic brain abnormalities, such as large meningiomas, primary brain tumors, or post-operative state
    \item Cases in which neuroradiologist JNS (2 years of experience) could not readily identify metastases after downsampling of images to 64 $\times$ 64 pixels (for reasons described below)
\end{itemize}

For metastasis-positive patients, the number of metastases ranged from one to over thirty / "innumerable." Metastases could be infratentorial and/or supratentorial in location. 

Within these exclusion criteria, we randomly selected 80 normal and 71 metastasis-containing patients / scans. These were initially selected based on the reports but verified on imaging review by neuroradiologist JNS. We partitioned these into a training set and separate testing set. 40 of the normal and 50 of the tumor-containing images were set aside for training for a total of 90 training images. 40 normal and 21 tumor-containing images were apportioned for testing,  a total of 61 testing set images. We allotted more tumor-containing images for training because, after the required downsampling to 64 $\times$ 64 pixels of the image slices (discussed below), the appearance of metastases was, though visible, less obvious than in the uncompressed images. 

Additionally, for the natural language processing portion of the study (discussed below), we extracted a separate set of only patient reports. We selected 45 reports from the normal data set and 45 reports from the metastasis-containing data set. Again, these reports were for patients not included in our image analysis case selection, and were selected randomly assuming that the diagnosis of no metastasis vs. metastasis was evident from inspection of the report impression. In order to include a broad range of language styles, we used images / reports that had been dictated by 18 neuroradiologists at our institution. In terms of patient demographics, the cohort was 61$\%$ male, 39$\%$ female. Racial / ethnic breakdown was: 84$\%$ White, 5$\%$ Asian, 6$\%$ Black / African American and 4$\%$ Hispanic. The three most common primary sites of disease were breast (24$\%$), skin (21$\%$) and lung (17$\%$). 

\subsection*{Part 1: Automated label extraction using natural language processing}

\subsubsection*{Attention-based string encoders, BERT, RoBERTa, and SBERT}

We endeavored to extract labels automatically from radiology reports. In order to do so, we employed a form of the powerful language representation model known as Bidirectional Encoder Representations from Transformers (BERT) \cite{devlin2018bert}. BERT uses a transformer model for encoding and decoding; here we use only the encoding portion. BERT's word encoding is essentially a vector that represents all aspects of a word. In addition to representing meaning(s) and connotations of words, the encoding is instilled with information about the word based on contexts in which it appears. Namely, positional encoding and the attention mechanism \cite{vaswani2017attention} imbue the ultimate word encodings with information about other words appearing in the same string (e.g., sentence or paragraph), including the order in which each word appears \cite{jawahar2019does}. BERT is trained on a very large corpus of written text using masked language prediction and next sentence prediction.

Robustly Optimized BERT Approach (RoBERTa) \cite{liu2019roberta} like BERT is an attention-based word embedding algorithm. It has nearly identical network architecture to BERT. However, it was trained only for masked language prediction, has more trainable parameters, and was trained on a larger corpus of text. In fact, RoBERTa comes pre-trained on 160 GB of text, including all of English Language Wikipedia and BookCorpus, itself containing over 11,000 books. As such, RoBERTa has demonstrated superior performance compared to BERT on a wide array of NLP tasks. As such, we use RoBERTa in this project. 

BERT/RoBERTa were intended to represent words for sequence-to-sequence tasks such as translation. However, they were not designed for representational learning of sentences, phrases, or paragraphs. Since we sought to capture the overall sentiment of a report impression section, we used required a mechanism to encode vectors with such information. Sentence BERT (SBERT) \cite{reimers2019sentence} is an adaption of BERT/RoBERTa that is more efficient at extracting the sentiment of whole sentences. It combines the word encodings from BERT/RoBERTa via dense interpolation / pooling, which maintains word meanings while minimizing representational vector size. We downloaded the pre-trained version of RoBERTa/SBERT via the the package sentence-transformers 0.3.0.

\subsubsection*{Application of SBERT with pre-trained weights and transfer-learning of SBERT to radiology reports}

As mentioned above, RoBERTa generated the word encodings, stacked into matrix form; we used the pretrained parameters from RoBERTa. Then the SBERT dense pooling of these provided matrices the rows of which represented encoded report impressions. Applying SBERT with only the pre-trained RoBERTa parameters gave 100 percent accuracy for normal images and 64 percent accuracy with brain metastases. We set out to improve this accuracy by transfer learning specific to radiology reports. 

To avoid mixing up the report impressions from the images we were analyzing for classification with the impressions on which we transfer-learned SBERT, we used a separate set of 90 impressions for the latter task. Half of these were for normal brain images and the other half were for metastasis-containing brain MRIs. 

In order to further train, i.e., do transfer learning with SBERT on our radiological report impression training set, we created all possible combinations of impressions, labeling these as the same or different meaning / category. Since this amounts to choosing all unique pairs out of a set of 90 elements (90 choose 2), the total number of generated pairs is given by: 
\begin{equation}\label{num_imp_pairs}
    N_{\text{pairs}} = {90 \choose 2} = 4,005
\end{equation}
Hence, we were able to generate a large training set from a relatively modest number of report impressions. Training SBERT on our impressions took 57 minutes and 22 seconds. After transfer learning on the radiology-specific jargon and prose of the reports, SBERT achieved 100$\%$ accuracy for both normal and metastasis-containing cases.

We can state this result a bit more formally if we denote by $l_{\text{radiologist}}$ the consensus class label between original reporting neuroradiologist and researching neuroradiologist (JNS, with 2 years of experience). We can also denote the class label predicted by the transfer-learned SBERT as $l_{\text{SBERT}}$. Then, denoting the size of the entire set of training and testing set images as $N_{\text{total}}$, given the aforementioned 100$\%$ accuracy of SBERT further trained on our radiology report corpus, we have
\begin{equation}\label{labels_NLP}
    \{ l_{\text{radiologist}} \}_{i=1}^{N_{\text{total}}}=\{ l_{\text{SBERT}} \}_{i=1}^{N_{\text{total}}}.
\end{equation}
Then, given their equivalence, we may henceforth refer to the true classes label for image $\mathcal{M}$ as $l_{\mathcal{M}}$.

\subsection*{Part 2: Reinforcement learning using labels from Part 1 to classify 3D images into normal and tumor-containing}

\subsubsection*{Reinforcement learning environment, definition of states, actions, and rewards}

As in all RL formulations, we must define our environment, itself defined by the agent, possible states, actions, and rewards.

\subsubsection*{Agent}

The agent in any RL formulation is an intuitive but formally murky concept. Essentially, the agent performs actions in various states and receives certain rewards. The goal of the agent is to act according to a policy (prescribed set of actions) that maximizes the total expected cumulative reward received. Hence, our goal, which aligns with that of the agent, is to learn the optimal policy for the agent to follow. 

\subsubsection*{States}

Here, the state is defined by an image volume plus a scalar quantity that indicates whether the previous class prediction (normal vs. tumor-containing) was correct. We denote this prediction correctness by $\text{pred}\textunderscore{\text{corr}}$. It is defined by 
\begin{equation}\label{corr_pred_defn_1}
    \text{pred}\textunderscore{\text{corr}} = \begin{cases}
        1 \text{, if prediction is correct} \\
        0 \text{, if prediction is wrong}  \text{.}
    \end{cases}
\end{equation}
Or equivalently,
\begin{equation}\label{corr_pred_defn_2}
    \text{pred}\textunderscore{\text{corr}} \equiv \delta_{a,l_{\mathcal{M}}},
\end{equation}
where $\delta$ is the Kronecker delta function.
Hence, we can write the state $s$ more formally as 
\begin{equation}\label{state_formal}
    s = \{ \mathcal{M}, \text{pred}\textunderscore{\text{corr}} \},
\end{equation}
where $\mathcal{M}$ is the $64 \times 64 \times 36$-voxel matrix representing the image volume. $\mathcal{M}$ is a $T1$-weighted post-contrast image volume with slice thickness of 5 mm. In order to make all matrices the same size, and to make the 3D convolutional backpropagation more computationally tractable, we downsize in the $x,y$-plane from original image size ($512 \times 512$ pixels for some images or $256 \times 256$ pixels for others) down to $64 \times 64$ pixels. Additionally, having different numbers of slices for the various images (ranging from 28 to 36), we make the $z$-axes of equal length by adding zero matrices of size $64 \times 64$ pixels to the caudal part of the images, so that all have $z$-dimension of 36. This is essentially providing padding in the z-direction.

\subsubsection*{Actions}

The two possible actions, $a_1$ and $a_2$, are simply the prediction of whether image $\mathcal{M}$ is normal (in the sense of no tumor present) or tumor-containing; i.e., the actions $\mathcal{A}$ are given by,
\begin{equation}
    \mathcal{A} = \begin{pmatrix} a_1 \\ a_2 \end{pmatrix} = \begin{pmatrix} 0 \\ 1 \end{pmatrix} = \begin{pmatrix} \text{predict normal} \\ \text{predict tumor} \end{pmatrix} \text{.}
\label{action_eqn}
\end{equation}

\subsubsection*{Rewards}

In order to encourage correct class predictions, we employ a reward that incentivizes correct predictions. Hence, we provide a reward of $+1$ if the prediction is correct and penalize with reward of $-1$ for a wrong prediction. We can define the reward $r$ in terms of prediction correctness, $\text{pred}\textunderscore\text{corr}$:
    \begin{equation}\label{rewards}
    r = \begin{cases}
        -1 \text{, if $\text{pred}\textunderscore\text{corr} = 0$} \\
        +1 \text{, if $\text{pred}\textunderscore\text{corr} = 1$}  \text{.}
    \end{cases}
\end{equation}

\subsubsection*{Action-value function ($Q$)}

A fundamentally important quantity in RL is that of a value function, in our case taking the form of the action-value function, denoted by $Q(s,a)$. $Q(s,a)$ essentially tells us the expected total cumulative reward that the agent would receive by selecting a particular action given a particular state, then acting in an optimal (on-policy) manner afterwards until the end of the current episode. An episode is defined as a number of states after which the agent begins anew in a different initial state. Restated, the action-value function represents the expected cumulative reward for taking action $a$ while in state $s$, if upon taking action $a$, a given policy (set protocol for selecting actions) is pursued thereafter until the end of the episode. More formally, given a policy $\pi$, the corresponding action-value, $Q^{\pi}(s,a)$, is defined by:
\begin{equation}\label{Q_defn}
    Q^{\pi}(s,a)=E_{\pi}\{R_t \arrowvert s_t=s,a_t=a\}=E_{\pi}\{\sum_{k=0}^{\infty} \gamma^k r_{t+k+1} \arrowvert s_t=s,a_t=a \} \text{,}
\end{equation}
where $R_t$ is the total cumulative reward starting at time $t$ and $E_{\pi}\{R_t \arrowvert s_t=s,a_t=a\}$ is the expectation for $R_t$ upon selecting action $a$ in state $s$ and subsequently picking actions according to $\pi$. The discount factor $\gamma$ represents the trade-off between weighting immediate rewards ("instant gratification") with rewards later on ("delayed gratification"). We set it equal to 0.99, as we have in prior work. 

The action-value function is critically important, because by maximizing this quantity, we can ultimately reach an optimal policy that produces a desired behavior, in this case correctly predicting image class.

\subsubsection*{Deep-$Q$ Network}

\begin{figure}
\centering
\includegraphics[width=12cm,height=9cm]{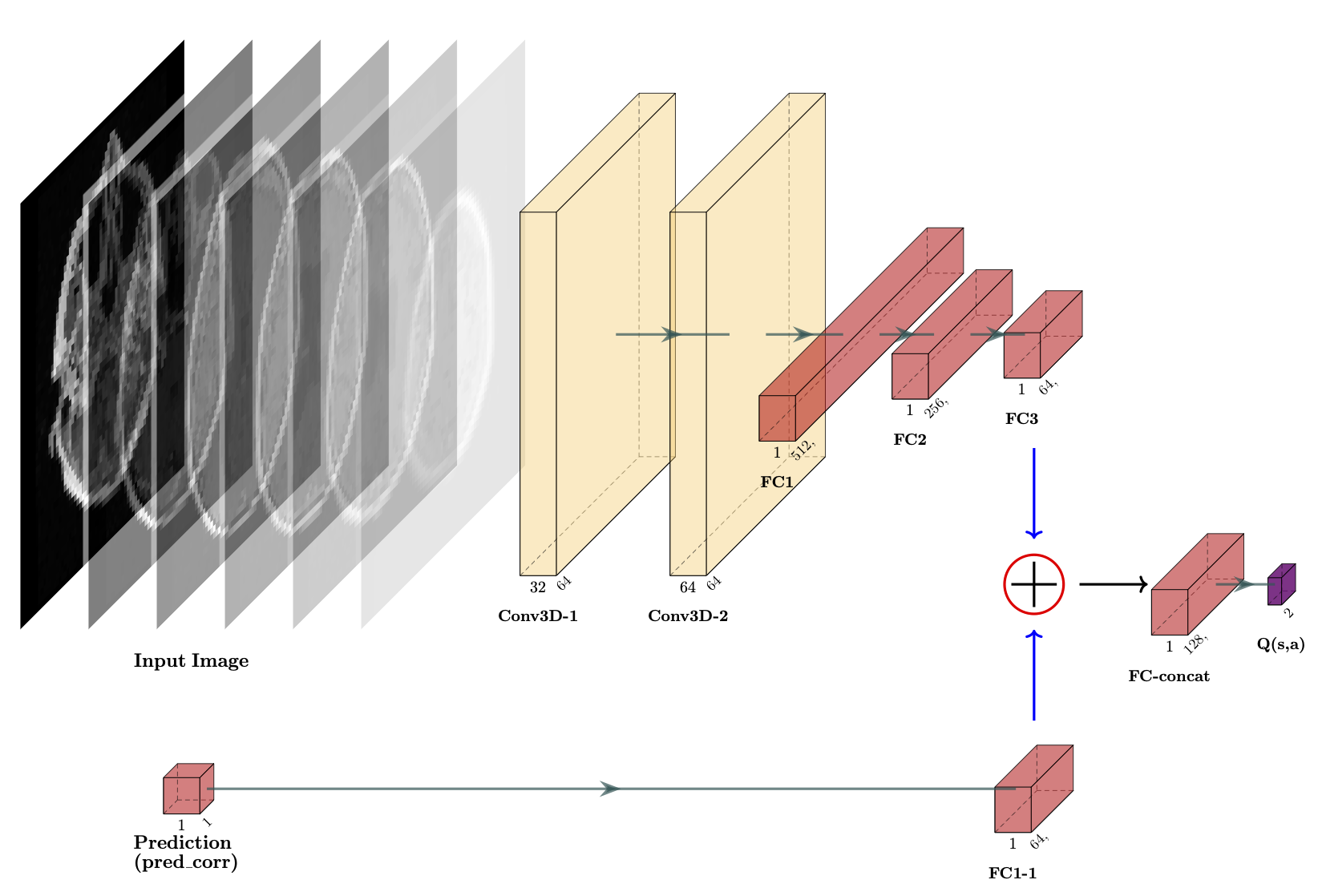}
\caption{ Schematic of the Deep $Q$ Network architecture.  }
\label{fig:dqn_arch}
\end{figure}

In order to predict the actions our agent will take, we use the Deep-$Q$ Network (DQN), as we have in prior work \cite{stember2020deep,stember2021deep,stember2020unsupervised,stember2020reinforcement}. The architecture of our DQN is illustrated in Figure \ref{fig:dqn_arch}. The image volume undergoes 3D convolutions, then the last convolutional layer is flattened and passed to a succession of fully connected layers. 

In a separate, parallel pathway, $\text{pred}\textunderscore{\text{corr}}$ is passed to a flattened layer, which is then concatenated with the last fully connected layer of the image volume network branch. This concatenated layer is then connected to a two-node output. 

The two-node output represents the two possible action-values, $Q(s,a_1)$ and $Q(s,a_2)$, that result from taking actions $a_1$ and $a_2$, respectively, from state $s$.  Again, since we wish to maximize the total cumulative reward, we should maximize $Q(s,a)$. We do so simply by selecting the $\text{argmax}_a \left( Q \right)$, thereby selecting the action that maximizes expected cumulative reward. 

We should note at this point that selecting $\text{argmax}_a \left( Q \right)$ is an "on-policy" action selection. As we will see, and have described in prior work \cite{stember2020deep,stember2021deep,stember2020unsupervised,stember2020reinforcement}, we also need to experiment with random action selections to explore and learn about the environment. This is called off-policy behavior. 

Indeed, an initial conundrum / catch-22 is that we wish to train our network to approximate the optimal policy's $Q$ function, but at first we have no idea what that optimal policy is. We can only start to learn it in pieces by sampling from the environment via off-policy exploration and Temporal Difference Learning. 

\subsubsection*{Temporal Difference Learning}

While the DQN allows us to select actions for our agent to take, we need to learn the best policy via the process of taking actions and receiving rewards. The rewards in particular tell us about our environment. We have to store these "experiences" of the agent in order to better understand the environment. By doing so in tandem with the use of DQN, we can train the parameters of the DQN to build it into a reliable approximator for the action-value function $Q(s,a)$. However, the optimal policy, and thus best possible action-value function, is a moving target that we approach through sampling. 

Then, using the DQN $Q$ function approximator for an ever-improving policy's $Q$ function, we can better explore and ultimately exploit what our agent knows about the environment to sample the environment more efficiently. The end result of this process is that DQN becomes a better and better approximator for not only a $Q(s,a)$ function, but the \textit{optimal} $Q(s,a)$ function, denoted by $Q^{\star}$.

In general, an episode of training proceeds as follows:

We select one of the training set images at random. The initial predicted class (normal or tumor-containing) is guessed at random. The initial value for $\text{pred}\textunderscore\text{corr}$ is set to zero as a default. At this point, an action is taken, i.e., a prediction as to the image class is made. As before, this is done in a manner that initially is more random, in order to emphasize exploration of the environment. As the agent learns about the environment and the best policy to follow, the degree of randomness decreases and the agent increasingly chooses the optimal action predicted by the DQN, i.e., the $\text{argmax}_a\{Q(a=0),Q(a=1)\}$. This is called the epsilon-greedy algorithm, and it is described in more detail in our prior RL works \cite{stember2020deep,stember2021deep,stember2020unsupervised,stember2020reinforcement}. We use similar parameters regarding this part of the approach:
\begin{itemize}
    \item $\epsilon = 0.7$
    \item $\epsilon_{min} = 1 \times 10^{-4}$
    \item $\Delta \epsilon = 1 \times 10^{-4}$,
\end{itemize}
where $\epsilon$ is the initial exploration rate, which decreases during training at a rate of $\Delta \epsilon$ down to a minimum value of $\epsilon_{min}$. 

Once we take the action from the prior step, this will bring us, at time $t$, from state $s_t$ to $s_{t+1}$ and will provide reward $r_t$. We can store all of these values in a tuple called a transition. As described in our prior RL work \cite{stember2020deep,stember2021deep,stember2020unsupervised,stember2020reinforcement}, we store these up to a maximum memory size called the replay memory buffer. Here, in order to conserve space and be able to save more transitions in the replay memory buffer, we store state information solely based on the training image number and correctness of the prior and just-made predictions, $\text{pred}\textunderscore\text{corr}_{t-1}$ and $\text{pred}\textunderscore\text{corr}_t$, respectively. We also store the index of the training image, $I_{\text{image}}$, which is all we need to retrieve the actual image volume $\mathcal{M}$ and class label $l_{\mathcal{M}}$. As such, the transition at time t, denoted by $\mathcal{T}_t$, is $\mathcal{T}_t=(\text{pred}\textunderscore\text{corr}_{t-1},I_{\text{image}},a_t,r_t,\text{pred}\textunderscore\text{corr}_t)$. Then the training batch for the DQN is selected randomly from the replay memory buffer, consisting of 15,000 transitions, stored as rows in a transition matrix. 

\begin{figure}
\centering
\includegraphics[width=11cm,height=7cm]{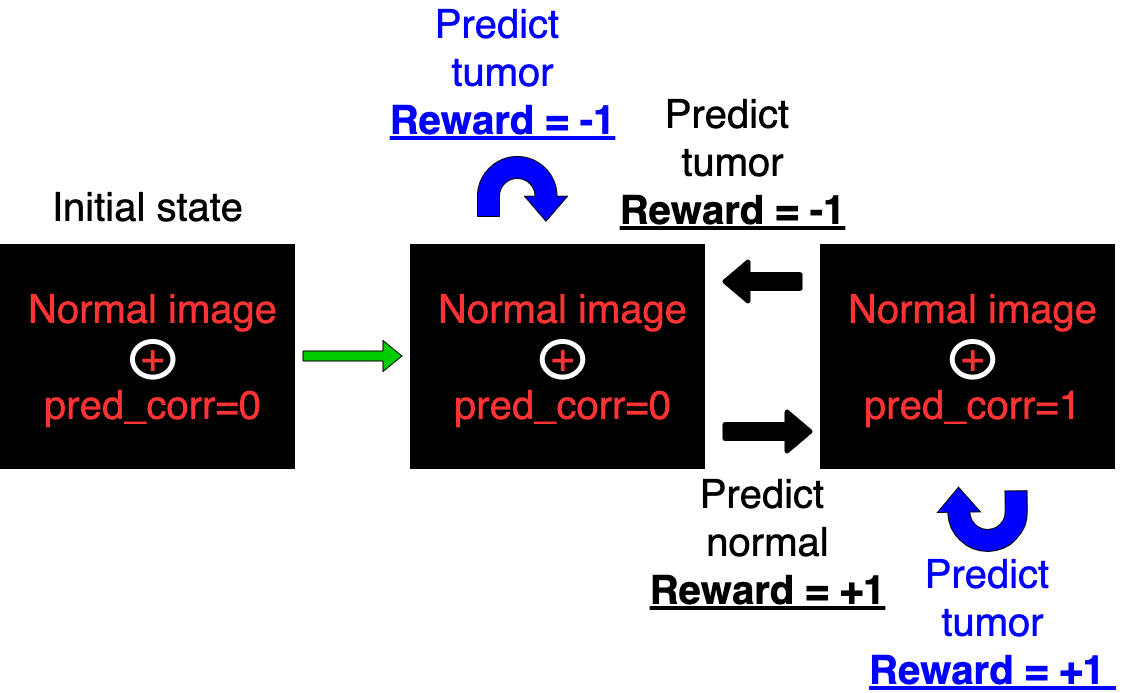}
\caption{ Markov Decision Process for a normal (no tumor) image.  }
\label{fig:MDP_normal}
\end{figure}

\begin{figure}
\centering
\includegraphics[width=11cm,height=7cm]{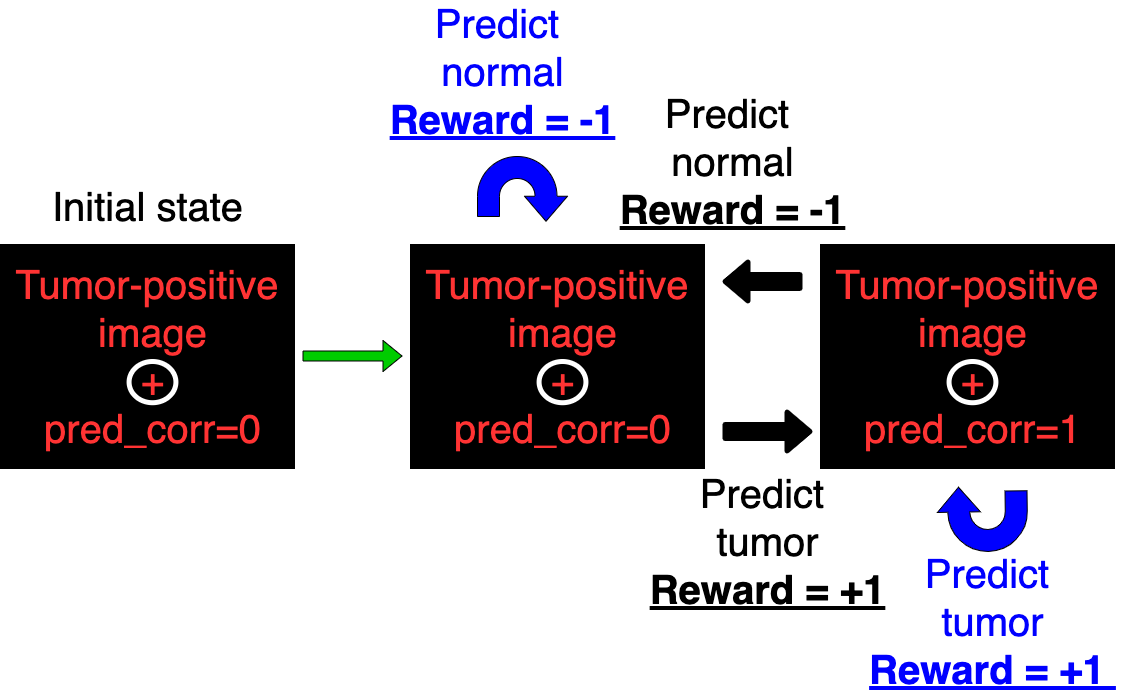}
\caption{ Markov Decision Process for a tumor-containing image.  }
\label{fig:MDP_tumor}
\end{figure}

In our prior work with 2D image slice classification \cite{stember2021deep}, we included information about the previous step as a color overlay on grayscale images. In order to save space by not storing the image itself with color overlay, we simply record the value for prediction correctness, $\text{pred}\textunderscore{\text{corr}}$ (0 if prior prediction was incorrect, 1 if it was correct). The default starting value for $\text{pred}\textunderscore{\text{corr}}$ in the initial state is zero. Then, if the correct prediction (which here is just the action) is made, we have $\text{pred}\textunderscore{\text{corr}} = 1$; otherwise, it remains zero. This Markov Decision Process is illustrated in Figures \ref{fig:MDP_normal} and \ref{fig:MDP_tumor}.

The Markov Decision Process (MDP) as shown in Figures \ref{fig:MDP_normal} and \ref{fig:MDP_tumor} proceeds as five steps per episode of training. At each step, we acquire data that samples the environment and is used to train the DQN. Each step of the MDP produces pieces of data about the environment called transitions. 

The agent is provided a reward of $+1$ for taking the correct action / class prediction, and is penalized with a reward of $-1$ for a wrong action / prediction. This is also shown in Figures \ref{fig:MDP_normal} and \ref{fig:MDP_tumor}.

\subsubsection*{Training: Deep-$Q$ Network architecture, parameters, hyperparameters}

Having verified 100$\%$ accuracy for label extraction, we trained an RL approach based on deep $Q$ networks (DQNs) and TD(0) temporal difference learning, as in recent work classifying 2D image slices \cite{stember2021deep}. In order to use TD(0), we employed the five-step Markov Decision Process illustrated in Figures \ref{fig:MDP_normal} and \ref{fig:MDP_tumor}. 

As shown in Figure \ref{fig:dqn_arch}, for our DQN, we used a two-input architecture. On one branch, the image volumes, again of size 64 $\times$ 64 $\times$ 36 voxels, were fed into a 3D convolutioal neural network as input. The 3D convolutional kernals were of size 5 $\times$ 5 $\times$ 5. As per usual, the weights in the kernals were initially randomized as per the standard Glorot distribution. Stride and padding were set as $\left( 2,2,2 \right)$ and $\left( 1,1,1 \right)$, respectively. The first convolutional layer has 32 channels, followed by a 64-channel layer. ReLu activation is employed after each convolutional step. The second convolutional layer is then flattened and passed to a fully connected layer of size 512 nodes, itself connected to a 256- then 64-node fully connected layer. 

Separately, the scalar $\text{pred}\textunderscore\text{corr}$, defined in Equations \ref{corr_pred_defn_1} and \ref{corr_pred_defn_2}, is connected to a 64-node fully connected layer. This layer is then concatenated with the the 64-node fully connected layer from the image convolutions. In this manner, information about both the image volume and whether the previous prediction of image class was correct are both included in the DQN. 

The concatenated 128-node layer is finally connected to two output nodes, $Q(a=0)$ and $Q(a=1)$. Most of the hyperparameters we used for training are similar to or the same as in our prior RL work \cite{stember2020deep,stember2021deep,stember2020unsupervised,stember2020reinforcement}. 
\begin{itemize}
    \item Loss: mean squared error
    \item Adam optimization
    \item learning rate: $1 \times 10^{-4}$
    \item batch size: 24
    \item Training time: 145 episodes
\end{itemize}

For predictions on the testing set, we ran individually through all of the testing set images after each set of 10 training episodes. Testing consisted of one step of applying the DQN to input state consisting of testing set image and initial $\text{pred}\textunderscore\text{corr}=0$. Then, the $argmax_a (Q(s,a)) $ was computed, and this action was the predicted class. Total training time was roughly 10 hours. 

\subsubsection*{Supervised deep learning (SDL) classification for comparison}

To compare SDL and RL-based classification, we trained the same $90$ training set images with a CNN with architecture essentially identical to the convolutional branch of the DQN from Figure \ref{fig:dqn_arch}. The CNN also consisted of 3D convolutional layers followed by elu activation employing $5 \times 5 \times 5$ filters. As for the DQN, this was followed by three fully connected layers. The network outputs a single node, which is passed through a sigmoid activation function given the direct binary nature of the CNN's prediction (normal vs. tumor-containing). The loss used here is binary cross entropy. Other training hyperparameters were the same as for the DQN. The supervised CNN was trained for $100$ epochs. 

\section*{Results}

\subsection*{Part 1}

As stated above, transfer-learning of SBERT to radiology reports resulted in 100$\%$ accurate label predictions based solely on the set of reports corresponding to the images analyzed in Part 2. Hence, we were able to use these labels in Part 2. 

\subsection*{Part 2}
Supervised deep learning classification, after training, predicted only class normal for the testing set, for an accuracy of 66$\%$, just over random guess given the distribution of normal and tumor-containing testing set images. Hence, supervised deep learning was unable to discern a difference between normal and tumor-containing images. Essentially, the trained network was unable to distinguish between the two classes and defaulted to the "first" class (label of zero for normal images). 

RL achieved an accuracy of 92$\%$. This difference in accuracy between SDL and RL is statistically significant, with a $p$-value of $3.1 \times 10^{-5}$. This was computed using the The McNemar’s test, which accounts for frequencies of false positives and false negatives and is widely used to compare machine learning classifiers. \cite{dietterich1998approximate}.

\section*{Discussion}

\subsection*{Part 1: NLP automated label prediction}

We have shown that transfer learning applied to an attention-based language encoder can produce very good results when applied to the specific verbiage of oncologic neuroradiology. This can enable automated label extraction directly from the electronic medical record. When coupled with extraction the images, this can allow for partial or completely automated training of classification algorithms, with little-to-no requirements from the user / researcher.

Of note, when the model was applied with only pre-trained weights that had been trained on more general text corpora such as Wikipedia, whereas they were perfectly accurate for normal scan report impressions, they did worse for the cases of brain metastases. This probably arose from the fact that, in our experience, there are fewer ways for radiologists to communicate normalcy, for example "normal scan" or "no intracranial metastasis." The language is simple, clear and does not vary much between interpreters. However, reporting a positive finding like new or increased metastases involves many more possible scenarios and ways of phrasing the findings. The vernacular of radiologists becomes more important, and the variability between radiologist phrasing increases. Hence, these were the scans for which SBERT improved dramatically by training specifically on radiology report impressions.  

\subsection*{Part 2: Image classification}

We have extended prior work using RL for classification to fully 3D image volumes from our institution. The statistically significant improvement for RL's 92$\%$ accuracy vs. the 66$\%$ of SDL fits with all our recent results showing much better performance of RL when trained on small training sets. RL is able to generalize more by exploring the data in a more exhaustive way that mimics how humans learn, whereas SDL employs a memorization process that is fragile in that it does not generalize well to new situations.

Some important limitations apply. Due to the considerable computational cost of 3D convolutions, we had to downsize our images often fourfold, producing pronounced loss of spatial and contrast resolution. Future work will seek to speed up the training process and eschew the downsizing, noting that brain metastases are often rather small and subtle, and in actual deployment downsizing is not an option.

\section*{Conflicts of interest}

The authors have pursued a provisional patent based on the work described here.

\begin{figure}
\centering
\includegraphics[width=11cm,height=7cm]{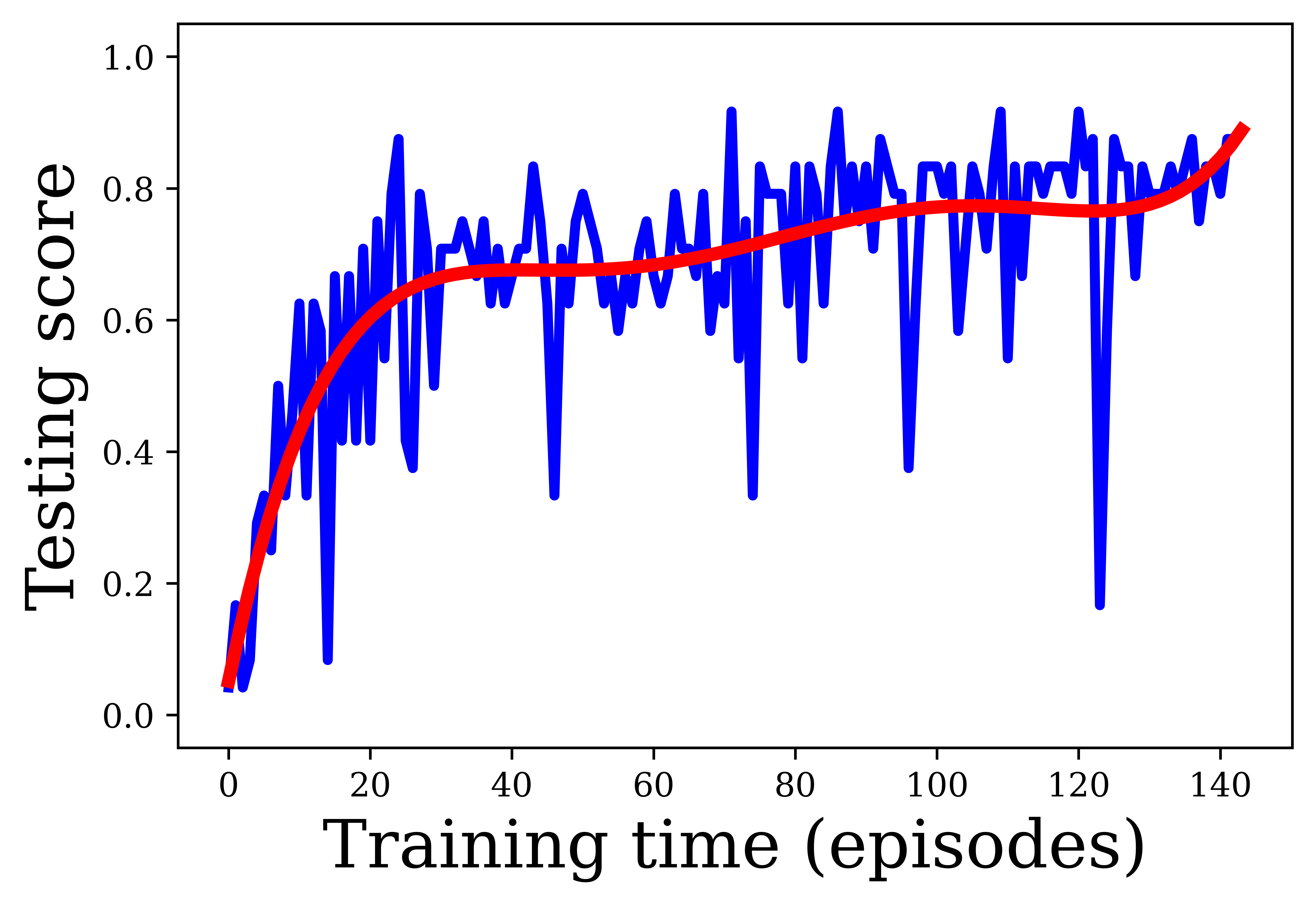}
\caption{ Testing set accuracy as a function of training time in episodes for reinforcement learning classification.  }
\label{fig:test_acc_during_training}
\end{figure}

\begin{figure}
\centering
\includegraphics[width=11cm,height=7cm]{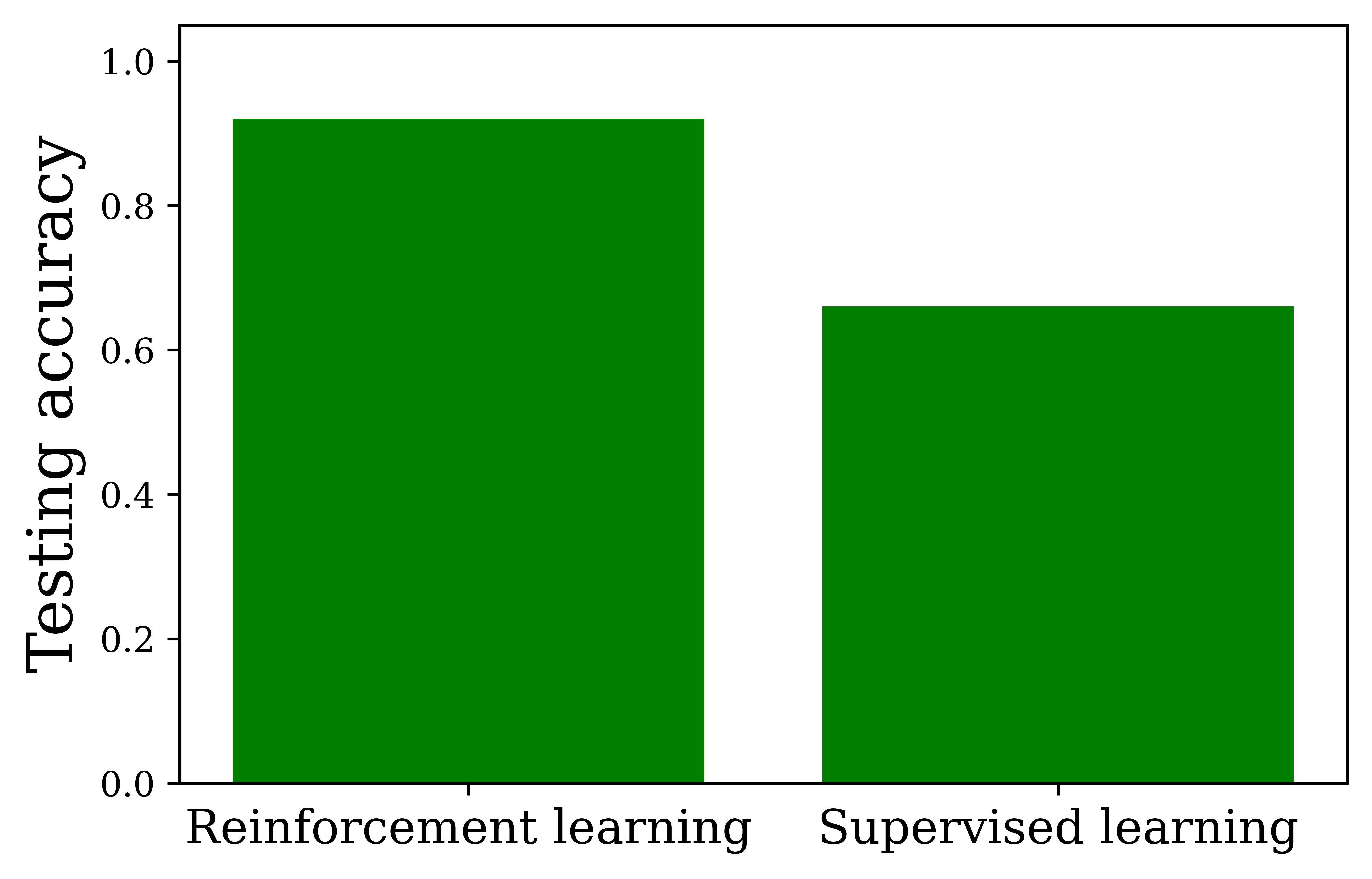}
\caption{ Comparison between reinforcement learning and supervised deep learning.  }
\label{fig:acc_comp}
\end{figure}

\printbibliography

@article{mcbee2018deep,
  title={Deep learning in radiology},
  author={McBee, Morgan P and Awan, Omer A and Colucci, Andrew T and Ghobadi, Comeron W and Kadom, Nadja and Kansagra, Akash P and Tridandapani, Srini and Auffermann, William F},
  journal={Academic radiology},
  volume={25},
  number={11},
  pages={1472--1480},
  year={2018},
  publisher={Elsevier}
}

@article{saba2019present,
  title={The present and future of deep learning in radiology},
  author={Saba, Luca and Biswas, Mainak and Kuppili, Venkatanareshbabu and Godia, Elisa Cuadrado and Suri, Harman S and Edla, Damodar Reddy and Omerzu, Toma{\v{z}} and Laird, John R and Khanna, Narendra N and Mavrogeni, Sophie and others},
  journal={European journal of radiology},
  volume={114},
  pages={14--24},
  year={2019},
  publisher={Elsevier}
}

@article{mazurowski2019deep,
  title={Deep learning in radiology: An overview of the concepts and a survey of the state of the art with focus on MRI},
  author={Mazurowski, Maciej A and Buda, Mateusz and Saha, Ashirbani and Bashir, Mustafa R},
  journal={Journal of magnetic resonance imaging},
  volume={49},
  number={4},
  pages={939--954},
  year={2019},
  publisher={Wiley Online Library}
}

@article{devlin2018bert,
  title={Bert: Pre-training of deep bidirectional transformers for language understanding},
  author={Devlin, Jacob and Chang, Ming-Wei and Lee, Kenton and Toutanova, Kristina},
  journal={arXiv preprint arXiv:1810.04805},
  year={2018}
}

@article{vaswani2017attention,
  title={Attention is all you need},
  author={Vaswani, Ashish and Shazeer, Noam and Parmar, Niki and Uszkoreit, Jakob and Jones, Llion and Gomez, Aidan N and Kaiser, Lukasz and Polosukhin, Illia},
  journal={arXiv preprint arXiv:1706.03762},
  year={2017}
}

@article{reimers2019sentence,
  title={Sentence-bert: Sentence embeddings using siamese bert-networks},
  author={Reimers, Nils and Gurevych, Iryna},
  journal={arXiv preprint arXiv:1908.10084},
  year={2019}
}

@article{liu2019roberta,
  title={Roberta: A robustly optimized bert pretraining approach},
  author={Liu, Yinhan and Ott, Myle and Goyal, Naman and Du, Jingfei and Joshi, Mandar and Chen, Danqi and Levy, Omer and Lewis, Mike and Zettlemoyer, Luke and Stoyanov, Veselin},
  journal={arXiv preprint arXiv:1907.11692},
  year={2019}
}

@inproceedings{jawahar2019does,
  title={What does BERT learn about the structure of language?},
  author={Jawahar, Ganesh and Sagot, Benot and Seddah, Djam},
  booktitle={ACL 2019-57th Annual Meeting of the Association for Computational Linguistics},
  year={2019}
}

@article{stember2020deep,
  title={Deep reinforcement learning to detect brain lesions on MRI: a proof-of-concept application of reinforcement learning to medical images},
  author={Stember, Joseph and Shalu, Hrithwik},
  journal={arXiv preprint arXiv:2008.02708},
  year={2020}
}

@article{stember2020reinforcement,
  title={Reinforcement learning using Deep Q Networks and Q learning accurately localizes brain tumors on MRI with very small training sets},
  author={Stember, Joseph N and Shalu, Hrithwik},
  journal={arXiv preprint arXiv:2010.10763},
  year={2020}
}

@article{stember2020unsupervised,
  title={Unsupervised deep clustering and reinforcement learning can accurately segment MRI brain tumors with very small training sets},
  author={Stember, Joseph and Shalu, Hrithwik},
  journal={arXiv preprint arXiv:2012.13321},
  year={2020}
}

@article{stember2021deep,
  title={Deep reinforcement learning-based image classification achieves perfect testing set accuracy for MRI brain tumors with a training set of only 30 images},
  author={Stember, Joseph and Shalu, Hrithwik},
  journal={arXiv preprint arXiv:2102.02895},
  year={2021}
}

@article{dietterich1998approximate,
  title={Approximate statistical tests for comparing supervised classification learning algorithms},
  author={Dietterich, Thomas G},
  journal={Neural computation},
  volume={10},
  number={7},
  pages={1895--1923},
  year={1998},
  publisher={MIT Press}
}

\end{document}